# An Adaptive Threshold for the Canny Edge Detection with Actor-Critic Algorithm

Keong-Hun Choi and Jong-Eun Ha

*Abstract*—Visual surveillance aims to perform robust foreground object detection regardless of the time and place. Object detection shows good results using only spatial information, but foreground object detection in visual surveillance requires proper temporal and spatial information processing. In deep learning-based foreground object detection algorithms, the detection ability is superior to classical background subtraction (BGS) algorithms in an environment similar to training. However, the performance is lower than that of the classical BGS algorithm in the environment different from training. This paper proposes a spatio-temporal fusion network (STFN) that could extract temporal and spatial information using a temporal network and a spatial network. We suggest a method using a semi-foreground map for stable training of the proposed STFN. The proposed algorithm shows excellent performance in an environment different from training, and we show it through experiments with various public datasets. Also, STFN can generate a compliant background image in a semi-supervised method, and it can operate in real-time on a desktop with GPU. The proposed method shows 11.28% and 18.33% higher FM than the latest deep learning method in the LASIESTA and SBI dataset, respectively.

## I. INTRODUCTION

EDGE information on images is useful for object detection, image segmentation, motion analysis, and 3D reconstruction [1, 2, 3, 4]. Traditional edge detection algorithms use a filter and statistical analysis. Recently deep learning has also been applied for detecting edges on images. The Canny algorithm proposed three decades ago is still widely used due to its good performance [5]. It requires the user to set three parameters related to the smoothing window size, and the remaining two are related to thresholds in the hysteresis process. We obtain very different edge images according to the values of three parameters, as shown in Fig. 1. In this paper, we propose an algorithm that can automatically determine values of three parameters suitable for the given images in the Canny algorithm.

In our previous work [6], we have proposed an algorithm that automatically determines values of two thresholds in the Canny algorithm using the Deep Q-Network (DQN) [7, 8]. We used a fixed value for a parameter related to the smoothing window size. In this paper, we propose an algorithm that can automatically determine the whole three parameters in the Canny algorithm. DQN is suitable for the discrete type of actions. When we extend our previous algorithm simply by adding additional parameters related to the size of the smoothing window, it can cause a problem due to many actions. Therefore, this paper copes with this problem by adopting an actor-critic algorithm [9, 10, 11] with continuous action. Also, we modify the edge evaluation network [6] to use both an original image and a corresponding edge image as input of the network to have improved evaluation of edge quality.

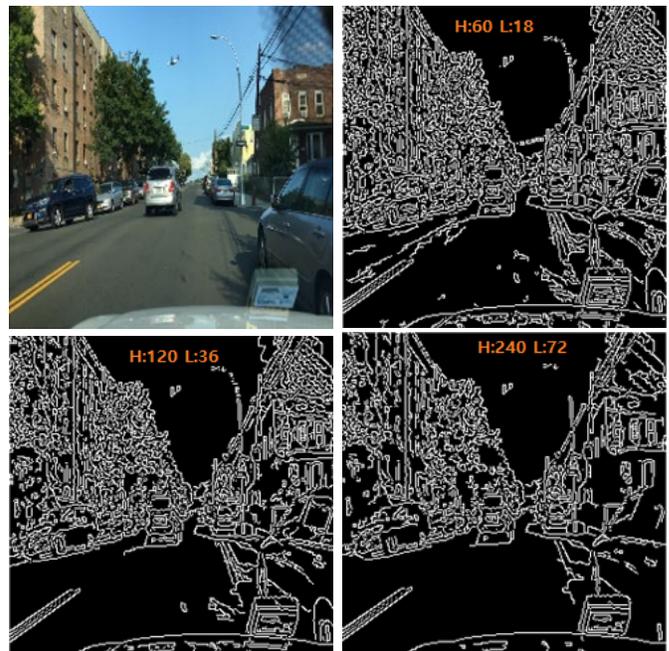

Fig. 1. The comparison of edge images according to different threshold values in the Canny algorithm.

A method based on supervised learning can also be considered for automatic threshold value setting. In the case of a supervised learning-based method, an end-to-end configuration is taken to extract the optimal edge image for the

This work was supported by the National Research Foundation of Korea(NRF) grant funded by the Korea government. (MSIT) (2020R1A2C1013335).

Keong-Hun Choi is with the Graduate School of Automotive Engineering, Seoul National University of Science and Technology, Seoul 01811, Korea (e-mail: temporary661@naver.com)

Jong-Eun Ha is with the Department of Mechanical and Automotive Engineering, Seoul National University of Science and Technology, Seoul 01811, Korea (e-mail: jeha@seoultech.ac.kr)

input image. It is necessary to secure the correct edge image for learning, which takes time. In general, it is known that supervised learning works well in cases similar to the learning environment but has poor performance in environments different from training. One way to adapt to a new environment is to retrain using an image of the corresponding domain. Also, this requires the preparation of ground truth labels.

In the case of the proposed method, a supervised learning-based CNN is used to specify a reward value. Still, when responding to a new environment, it is shown that performance improvement is possible through additional unsupervised learning without updating such the CNN. The method proposed in this paper has the advantage of improving performance by using only images from a new environment without requiring ground truth labels.

The contributions of the proposed algorithm are as follows.

(1) We present a method for automatically selecting all three threshold values that guarantee an excellent edge image quality with the Canny algorithm. We solve the problem caused by many combinations of three threshold values by estimating continuous action based on actor-critic.

(2) We present a reward model for stable learning of action network and critic network in the actor-critic algorithm. We configure the reward model by reflecting two terms. One is the output of edge evaluation CNN. The other is a constraint for preventing reversal of high and low thresholds. The edge evaluation CNN uses an original image and the edge image as inputs and produces a value for the quality of the edge.

(3) The proposed algorithm can improve the performance by using only images without ground truth labels in the new environment.

## II. Related works

Edge detection algorithms can be categorized into three groups: filter-based, learning-based, and recent deep learning-based algorithms.

Traditional filter-based algorithms [3] find edges by investigating dramatic changes in intensity, color, texture, etc.

Learning-based algorithms detect edges through supervised learning with hand-crafted features. Statistical Edges [12], Pb [13], and gPb [14] use features derived by careful manual design based on information theory. Early learning-based methods such as BEL [15], Multi-scale [16], Sketch Tokens [17], and Structured Edges [18] also heavily rely on features designed manually. Dollar et al. [18] detect structured edges by joint learning ground truth clustering and mapping image patches to clustered tokens. It showed state-of-the-art performance on the BSDS500 dataset until the advancement of deep learning-based algorithms.

Learning-based algorithms have the advantage that they can automatically generate edge images by reflecting structured information on images. But, their applicability is shown using only a small number of images compared with a large number of images used in deep learning.

Recent deep learning-based algorithms utilize features that are generated by convolutional neural networks (CNN) [19]. Bertasius et al. [20] use CNN to find features of candidate contour points. Xie et al. [21] propose holistically-nested edge detection (HED) that integrates the outputs from different intermediate layers with skip connections. Xu et al. [22] use a hierarchical model to find multi-scale features fused by a gated conditional random field. He et al. [23] propose a Bi-Directional Cascade Network (BDCN) structure to detect edges at different scales. They train the network using other labeled edges for each scale. They introduced Scale Enhancement Module (SEM), which utilizes dilated convolution to generate multi-scale features instead of using deeper CNNs or explicitly fusing multi-scale edge maps. Recent algorithms for edge detection focus on the accurate detection of object boundaries that can provide semantics cues for further processing, such as object detection, segmentation, and tracking.

Lu et al. [24] proposed an algorithm to select thresholds for the Canny algorithm using a histogram of the gradient image. Fang et al. [25] proposed an algorithm to choose a high threshold for the Canny algorithm using Otsu method [26]. They cannot select a low threshold. Huo et al. [27] proposed an algorithm to determine the Canny algorithm's high and low threshold. They choose a low threshold using a probability model. Lu et al. [28] adaptively select two thresholds using minimal meaningful gradient and maximal meaningless gradient magnitude assumption. Yitzhaky and Peli [29] proposed choosing the best edge parameters. First, they construct Estimated Ground Truth (EGT) with different detection results. Then, they determine the optimal parameter set using a Chi-square test. Medina-Carnicer et al. [30] proposed an algorithm for the unsupervised determination of hysteresis thresholds by fusing the advantages and disadvantages of two thresholding algorithms. They find the best hysteresis thresholds in a set of candidates. Mediana-Carnicer et al. [31] proposed a method to determine hysteresis thresholds of the Canny algorithm automatically. Therefore, it can be used as an unsupervised edge detector.

These traditional unsupervised methods have the advantage that they can be applied without a learning process. However, in the case of these methods, only evaluation results for a small number of images are provided. It is necessary to evaluate them using a lot of data, as in deep learning, to objectively evaluate these algorithms' performance.

Reinforcement learning [32] shows a good performance in temporal decision-making problems. In a typical reinforcement learning, an agent aims to learn a policy that gives maximum accumulated reward from an environment. Recently, integrating reinforcement learning and deep learning showed human-level control [33]. Deep Q-Networks (DQN) [7, 8] showed human-level control is possible on Atari games. Deep reinforcement learning offers impressive successes on various tasks such as playing the board game GO [34, 35, 36], object localization [37], region proposal [38], and visual tracking [39].

## III. proposed Method

This paper proposes a method for automatically selecting the three threshold values required for the Canny edge method according to the image, using the advantage actor-critic (A2C)

method [9,10,11].

Fig. 2 shows the resulting edge images according to the varying threshold values in the Canny edge. We notice that the difference in the resulting edge image is enormous according to the threshold value change. However, the resulting edge images are similar in the range of similar threshold values. Therefore, we assume that the resulting edge image would follow a normal distribution according to the threshold value. We could regard the threshold value for the Canny algorithm as action in deep reinforcement learning. Therefore, through deep reinforcement learning, we could automatically find appropriate threshold values that guarantee an excellent edge image by the Canny algorithm. Based on this assumption, we propose a method based on A2C [9, 10, 11] to automatically determine the three threshold values required in the Canny algorithm.

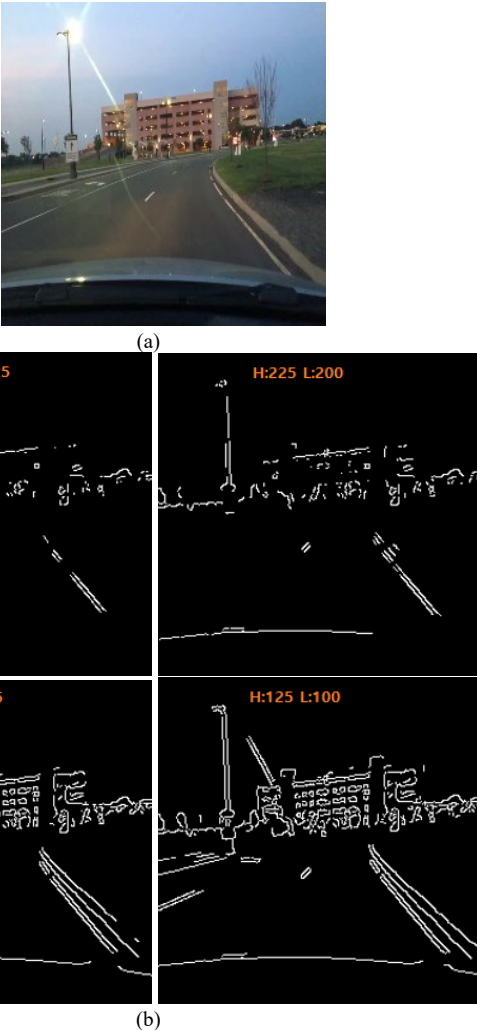

Fig. 2. The similarity in the resulting edge image according to the threshold values in the Canny algorithm (a) original image (b) resulting edge image.

### A. Overview of Advantage Actor-Critic (A2C) Algorithm

In the standard reinforcement learning setting, an agent interacts with an environment $\mathcal{E}$ over several discrete time steps. The agent selects an action $a_t$ at a state $s_t$ by policy $\pi$, where $\pi$ is a mapping from states $s_t$ to actions $a_t$. An action $a_t$ is chosen from some set of possible actions A. After the action, an agent gets to the next state $s_{t+1}$ and receives a scalar reward $r_t$. The process continues until the agent reaches a terminal state. The total accumulated rewards from time step $t$ is denoted return as follows.

$$R_t = \sum_{k=0}^{\infty} \gamma^k r_{t+k} \quad (\gamma \in (0,1]) \quad (1)$$

The agent wants to maximize the expected return from each state $s_t$. Policy-based model-free methods directly parameterize the policy $\pi(a|s;\theta)$ and find the parameters $\theta$ that maximize $E[R_t]$. The REINFORCE family of algorithms can be adopted [9]. Typical REINFORCE updates the policy parameters $\theta$ in the direction $\nabla_\theta log\pi(a_t|s_t;\theta)R_t$, which is an unbiased estimate of $\nabla_\theta E[R_t]$. It is possible to reduce the variance of this estimate while keeping it unbiased by subtracting a learned function of the state $b_t(s_t)$, known as a baseline [9], from return. The resulting gradient is as follows.

$$\nabla_\theta log\pi(a_t|s_t;\theta)(R_t - b_t(s_t)) \quad (2)$$

A learned function of the value function $V^\pi(s_t)$ is commonly used as the baseline, leading to a much lower variance estimate of the policy gradient. The advantage of action $a_t$ at state $s_t$ is defined as follows.

$$A(a_t, s_t) = Q(a_t, s_t) - V(s_t) \quad (3)$$

$Q^\pi(a_t, s_t) = E(R_t|s_t = s, a_t = a)$ is an action value corresponds to the expected return for selecting action a in state s following policy $\pi$. Since $R_t$ is an estimate of $Q^\pi(a_t, s_t)$ and $b_t$ is an estimate of $V^\pi(s_t)$, $A(a_t, s_t)$ can be used for $R_t - b_t(s_t)$ in Eq. (2). This approach can be viewed as an actor-critic architecture where the policy $\pi$ is the actor and the baseline $b_t$ is the critic [10].

If we use one-step return, the Eq. (2) can be represented as follows.

$$\nabla_\theta log\pi(a_t|s_t;\theta)\big(R_{t+1} + \gamma V(s_{t+1};\theta_v) - V(s_t;\theta_v)\big) \quad (4)$$

In advantage actor-critic algorithm, two networks of $\pi(a_t|s_t;\theta)$ and $V(s_t;\theta_v)$ are used.

### B. State Configuration

When we regard an original image as a state $s_t$, we obtain an edge image as the result of the action that corresponds to the three threshold values of the Canny algorithm. If such an edge image is considered as $s_{t+1}$, it may cause problems due to properties different from $s_t$. In the case of the value network $V(s_t;\theta_v)$, an image is used as an input, and when calculating $V(s_{t+1};\theta_v)$ in Equation (4), an edge image must be used as an input, which gives incorrect results.

In this study, to solve this problem, we use a randomly selected image among the images used for training rather than an edge image as $s_{t+1}$. In the case of a properly trained model, we can assume that appropriate actions are selected for randomly selected images. Therefore, the value evaluation for the randomly selected image would be similar to the value evaluation by considering future actions. Through this, we can apply the A2C algorithm to a computer vision problem in which an input image and an output image with different properties are present.

*C. Actor and Critic Network Configuration*

An actor network for choosing actions and a critic network for evaluating the accumulated return from the current state are required to train A2C. There are two types of configuring these networks in the A2C algorithm. The first method uses the same backbone and has two branch outputs for the actor and critic. The second method uses two different networks for the actor and critic. Choosing the first method with the same backbone gave inconsistent results, including divergence during training. Therefore, the actor and critic networks are divided into independent network structures in this paper. The A2C algorithm has the advantage of being able to output continuous action values compared to DQN. If the policy is assumed to be a normal distribution, it will have a continuous output as follows.

$$\pi(a_t|s_t;\theta) = \frac{1}{\sigma(s_t;\theta)\sqrt{2\pi}} exp\left(-\frac{(a_t-\mu(s_t;\theta))^2}{2\sigma(s_t;\theta)^2}\right) \quad (5)$$

$\mu(s_t;\theta)$ and $\sigma(s_t;\theta)$ is the mean and standard deviation of the Gaussian distribution.

Fig. 3 shows the configuration of the proposed actor network. We use an original image as the network's input, and the network's output is the mean and variance of three actions. We select threshold values by random selection from the Gaussian distribution. The actor network determines the mean and variance of the Gaussian distribution.

We extract features of the original image using a pre-trained CNN of ResNet50 structure. In the case of the fully connected (FC) layer that outputs the mean and variance, the training was performed by initializing the factor values at random. The distribution of the three threshold values follows a Gaussian distribution, and the range of values ranges from -1 to 1. We use tanh as the activation function of the last layer for the output of the mean. We use sigmoid as the activation function of the last layer for the output of the standard deviation.

The three output values of the actor network are continuous real values, and it is necessary to convert them to the threshold value of the Canny algorithm. We convert the output of the actor network into an integer from 0 to 500 for the high and low threshold. For the filter size of smoothing window, we convert the output of the actor network into an integer between 3 and 9.

Fig. 4 shows the configuration of the proposed critic network. The output of the network corresponds to $V(s_t)$. The input of the critic network is composed of an original image and action values. We randomly select the action values used as an input of the network from the normal distribution provided by the actor network.

We use action values as the input of the critic network since the critic network is a model that approximates the $V(s_t)$ value according to each action. The output of the actor network is the mean and variance of the Gaussian distribution, which is used for sampling three threshold values. Therefore, we use threshold values sampled from the Gaussian distribution as input of the critic network to obtain $V(s_t)$. Through this, we can guarantee that training would increase the probability of selecting a specific value from the normal distribution of the actor network.

Considering that the critical network is a model that approximates the evaluation score according to the action, we do not use an activation function in the output layer. In addition, we concatenate three action values with features from the pre-trained model in the middle of FC layer to prevent them from being reflected in a small proportion.

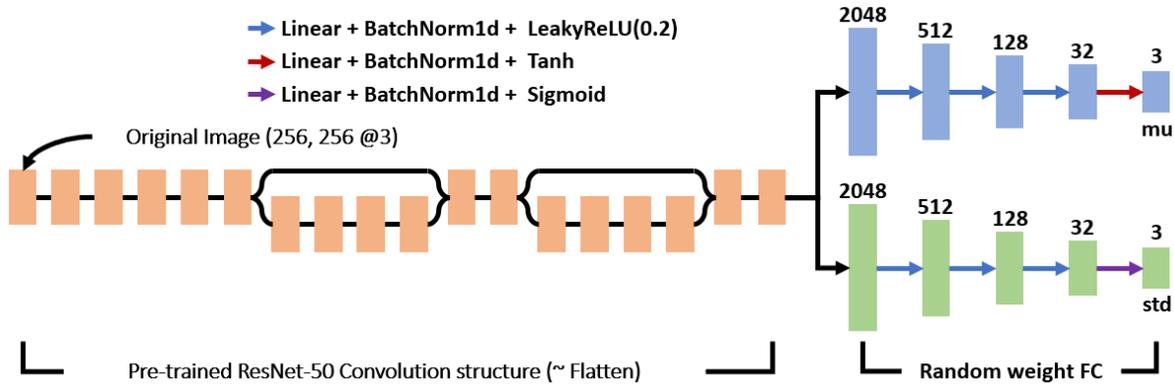

Fig. 3. The structure of the proposed actor network.

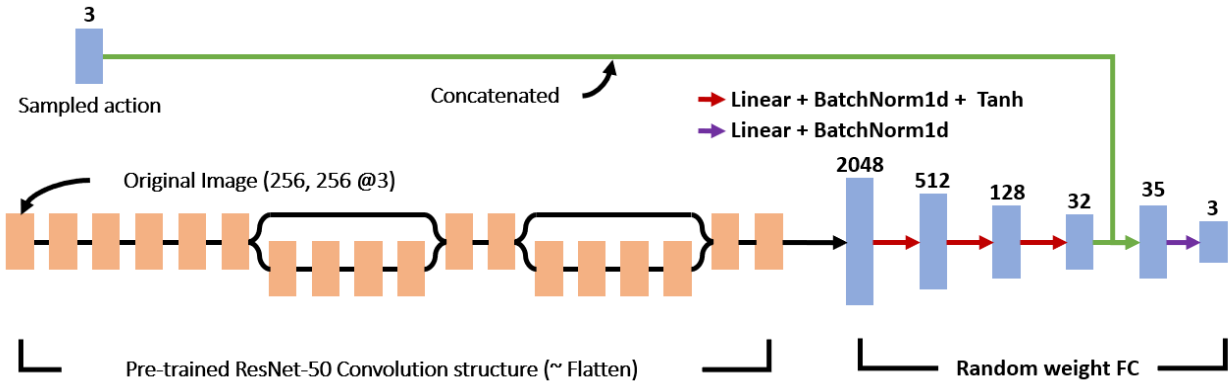

Fig. 4. The structure of the proposed critic network.

*D. Reward Computation*

In reinforcement learning for robot posture control, we can easily set rewards by simply checking straight or falling. However, in the case of the automatic selection of the three threshold values of the Canny edge to be solved in this paper, reward selection is not simple. Calculating the reward by evaluating the edge image to which the three selected threshold values are applied is necessary. To this end, in this paper, the reward is calculated using a supervised CNN trained. Like the disadvantages of the existing supervised learning, it requires a process to generate separate label data. It has a disadvantage in that generalization ability is poor in an environment that is not similar to the learning data.

An edge evaluation model was constructed that finally outputs the edge fit using the original image and the resulting edge image as inputs. Fig. 5 shows a model that outputs the fit of the edge image by inputting the original image and the edge image. The edge evaluation model uses original image and edge image as inputs and extracts feature values using the pre-trained CNN structure of ResNet-50. The extracted feature value goes through a fully connected layer to output the edge fit. The last layer of the fully connected layer uses the sigmoid activation function, and through this, the output value has a value between 0 and 1.

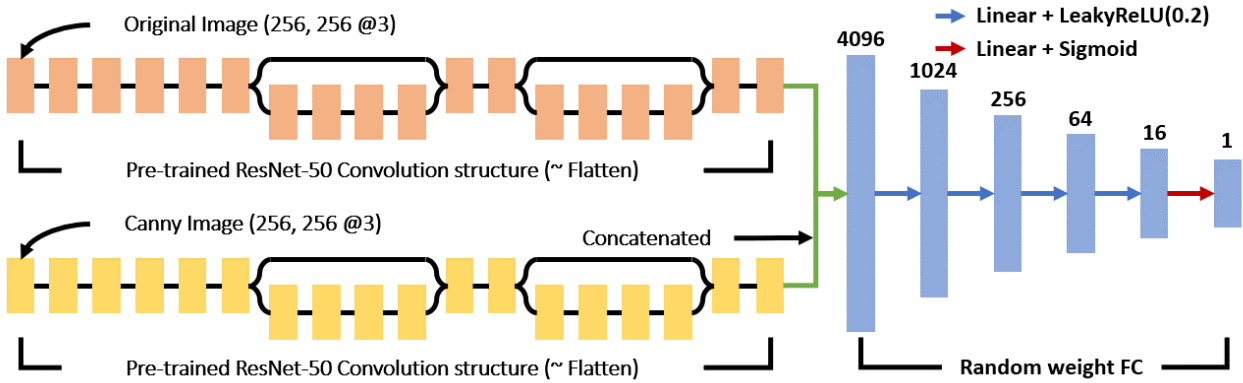

Fig 5. The structure of the proposed edge evaluation model.

Since the output of the proposed edge evaluation model has a value between 0 and 1, we always have positive output values. When only positive reward values are used, feedback on wrong actions is impossible in reinforcement learning [40]. In addition, it is necessary to take measures to suppress the occurrence of high and low threshold reversal.

Table I shows the reward configuration reflecting these two points. We make the reward value have a value in the range of positive and negative according to the output of the edge evaluation model. When the high and low threshold value reversal occurs, we set the reward to a negative value. We experimented with four cases with different reward values in the case of high and low threshold reversal. Reward values are negative but have different values in four cases.

TABLE I
REWARD VALUE CONFIGURATION BY REFLECTING ACTION RESULT AND EDGE EVALUATION RESULT

| action | edge evaluation | reward value | | | |
|---|---|---|---|---|---|
| | | case 1 | case 2 | case 3 | case 4 |
| high ≧ low | >= 0.999 | 1.0 | 1.0 | 1.0 | 1.0 |
| | >=0.500 | 0.5 | 0.5 | 0.5 | 0.5 |
| | < 0.500 | -0.5 | -0.5 | -0.5 | -0.5 |
| high < low | - | -1.0 | -5.0 | -10.0 | -50.0 |

## IV. EXPERIMENTAL RESULTS

We resize an image into 256(H)X256(W)X3(C). Each pixel is divided by 255, which results in a value between 0 and 1. The training was done using NVIDIA 3090 and Intel i9-10900. The Adam optimizer was used for training. The learning rate of the optimizer was 0.001 with eight batch sizes.

### A. Edge Evaluation Model

We used 600 images from BDD100K [41] to train the edge evaluation model of Fig. 5. We manually generated two negative and positive images from each image, and then we augmented data through geometric transformations. Finally, we used 20,000 images for training. We used the mean squared error (MSE) between the output result and the label data for the loss of the edge evaluation model. The accuracy is defined as follows.

$$ACC = (1 - |O_g - O_p|) \times 100 \tag{6}$$

$O_g$ is a ground-truth value where positive edge images have a value of 1 and negative edge images have a value of 0. $O_p$ is an output value of the edge evaluation model, and it has a value between 0 and 1.

Fig. 6 shows the training results, and we did the training for 14 epochs. We use the result of the 11th epoch, which is the case where the accuracy value using the validation data is the best.

Fig. 7 shows the results of the edge evaluation model using the original image and the edge image as inputs and the edge evaluation model [6] using only the edge image as inputs. The model using both the edge and original images provides improved results compared to using only the edge image as an input.

Images acquired at nighttime have low contrast. Therefore, it is challenging to extract edges. The previous edge evaluation model [6] has difficulty in those images, as shown in Fig. 7. It gives a high evaluation for a low-quality edge image. The proposed edge evaluation model gives a correct assessment even in this case. Improved evaluation is possible by using the original image and the edge image as the input of the evaluation model compared to the previous evaluation model [6], which only uses an edge image.

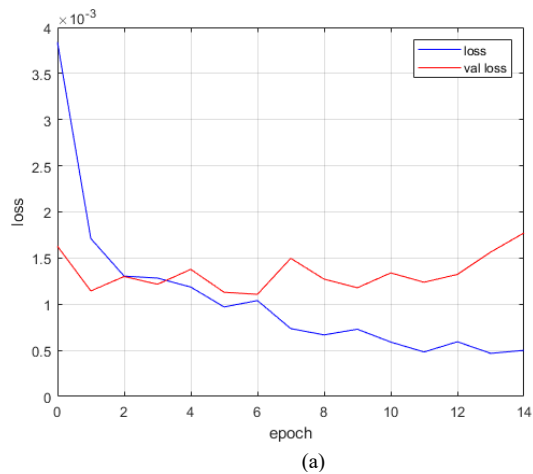

(a)

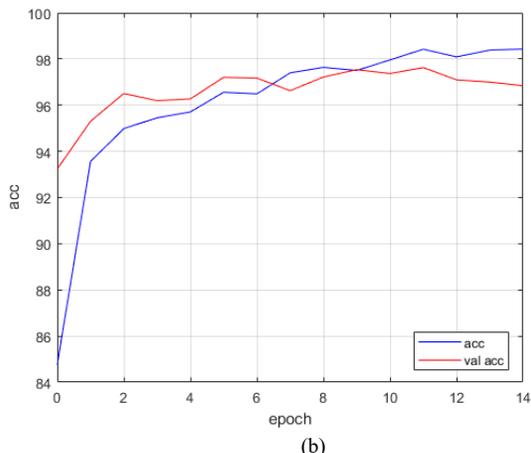

(b)

Fig. 6. The variation of loss and accuracy during training (a) loss (b) accuracy.

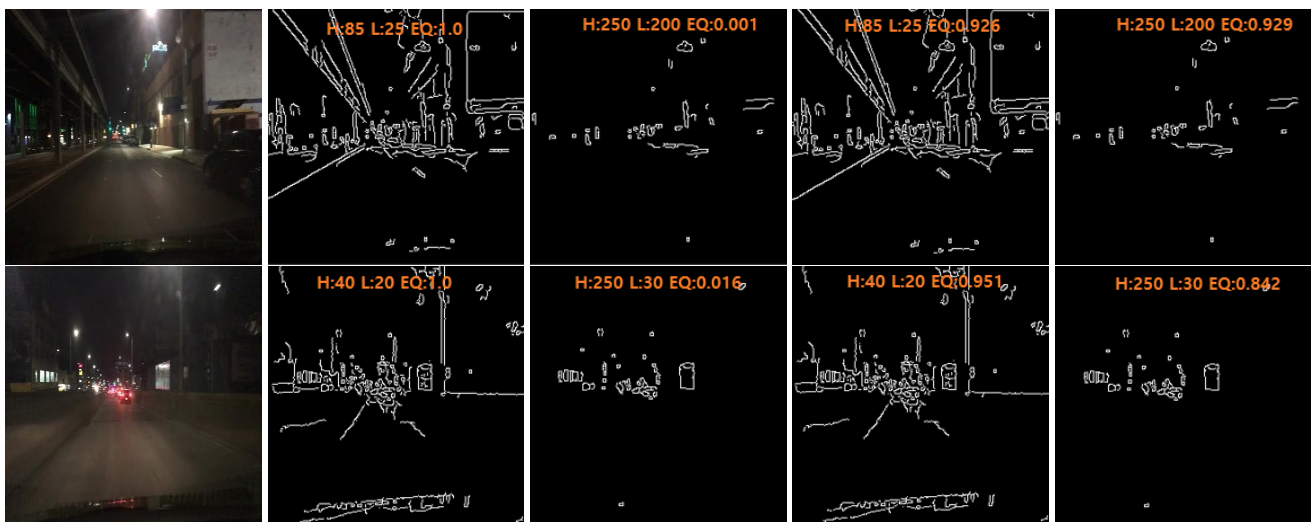

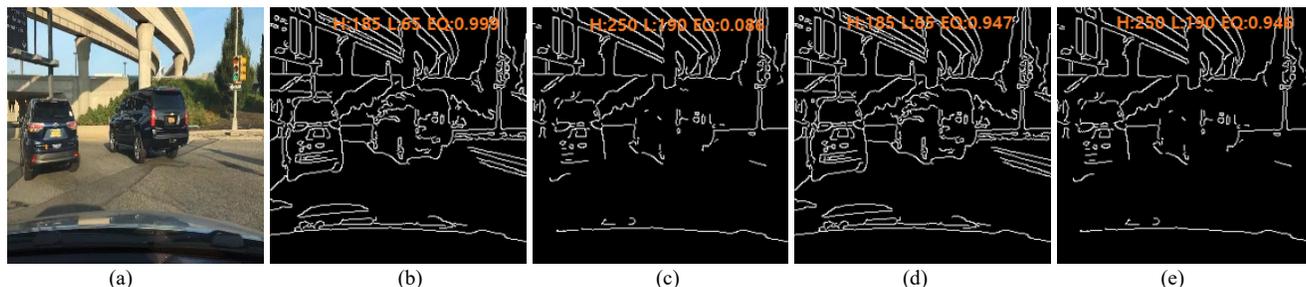

(a) (b) (c) (d) (e)

Fig. 7. The comparison result of the proposed edge evaluation model and model in [6] (a) original image (b) proposed algorithm for a good edge image (c) proposed algorithm for a bad edge image (c) [6] for a good edge image (d) [6] for a bad edge image (EQ: stand for edge equality).

## B. Result of the Proposed Method

We obtain the three threshold values needed for the Canny edge algorithm from the output of the actor network of Fig. 3. We used 50,000 images in BDD100K[41] for the training. We use three items for the evaluation of the training of the network. We use the change of the variance of the output of the actor network, the reversal ratio of the high and low threshold values, and the average of the output of the edge evaluation network.

The actor network can specify threshold values suitable for the input image in a narrow range when training is adequately performed. Therefore, when training is done in an appropriate direction, the size of variance, one of the two outputs of the actor network in Fig. 3, will gradually output a smaller value.

Fig. 8 shows the change of variance of three actions according to four cases of reward configuration during training. In Fig. 8, we notice that the variance's magnitude is continuously decreasing for all cases. It indicates that the training is proceeding correctly.

The variance of smoothing window size has a relatively large value, while the variance of high and low thresholds falls below 0.3. We convert the value of smoothing window size into a range between [3, 9] and high and low thresholds between [0, 500]. We can conclude that the relatively high variance of smoothing window size is acceptable considering the difference in conversion range. Among the four reward cases in Table I, case 3 shows the smallest variance. Through this, case 3 can be considered the most suitable reward model.

Fig. 9 shows the change in reward values of the four reward models in Table I during training. The average of the reward values of the last 80 cases for each trial is shown. We notice that training is being appropriately performed by continuously improving the average reward value until about 1.6 million steps. Finally, the average of the reward values has values larger than 0.5. We can conclude that the edge evaluation model obtained values more than 0.5 in most images used for training by referring to the reward value configuration in Table 1. The edge evaluation network will have a high output value when the actor network is appropriately trained.

Fig. 10 shows the change of output value of the edge evaluation model during training. We can notice that it gives a similar tendency to Fig. 9. Also, we indirectly convince that the proposed algorithm is well trained in a direction that offers appropriate threshold values for the Canny algorithm.

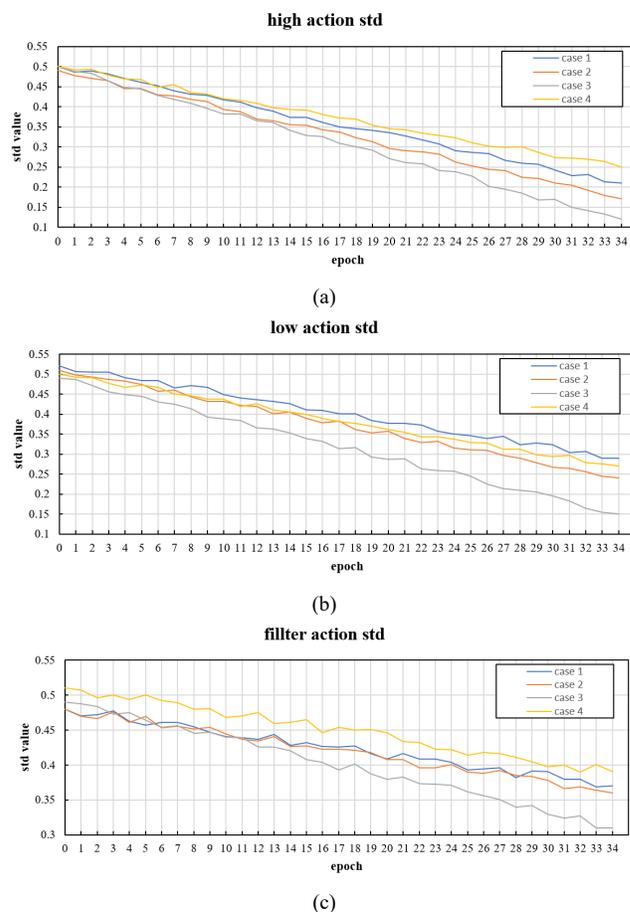

Fig. 8. The change of variance of three actions according to the four reward cases during training (a) high threshold (b) low threshold (c) smoothing window size.

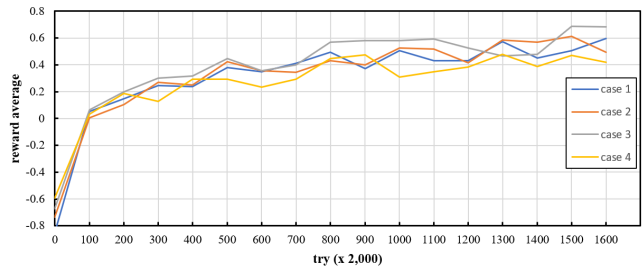

Fig. 9. The variation of the mean reward of the last 80 samples during training.

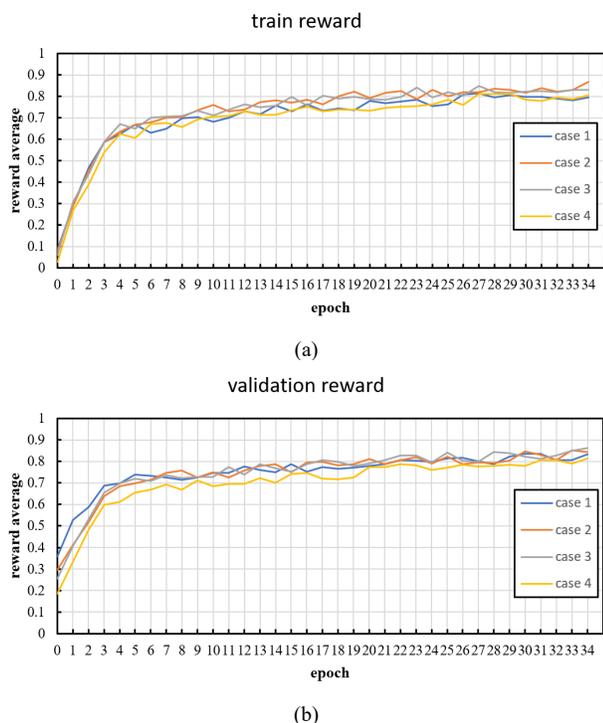

Fig. 10. The variation of quality value of the edge evaluation model during training (a) training set (b) validation set.

Table II shows the comparison results according to the edge evaluation model and compensation model of the existing DQN method [6] and the proposed method. Statistics in Table 2 are obtained from the result of 20,000 images.

The proposed model has a structure in which high and low threshold values are distinguished and output. Also, in the case of the reward used for training, when the low threshold value is greater than the high threshold value, training was carried out using a negative reward value. Therefore, if the model is properly trained, the high threshold should always be greater than the low threshold. Therefore, it is possible to determine the learning tendency of the model by checking the occurrence of the reversal of the high threshold value and the low threshold value.

Table II shows that it is necessary to give a negative reward when a reversal of a high threshold value and a low threshold value occurs. In addition, it can be seen that in the case of inversion of the threshold value, it is necessary to maintain a particular value rather than increasing the size of the negative reward indefinitely. The DQN method [6] has the advantage of no reversal between high and low thresholds. But, it gives a reward value smaller than that of the proposed method.

The proposed algorithm gives superior results than the DQN method [6] in all four cases. But 96.7% among the 20,000 images were obtained without reversing high and low thresholds in the best case 3. In contrast, in the case of the DQN method, the threshold value reversal does not occur in all cases.

The reversal of high and low thresholds in the proposed algorithm is inherently connected to the procedure that randomly selects the mean and variance from the output of the actor network. There is a radical possibility of reversing high and low threshold values during sampling even if the proposed network is well trained. It requires further research to solve this problem. In contrast, in the DQN method, it is structurally possible to configure the high threshold always to have a bigger value than the low threshold.

Fig. 11 shows the comparison results of edge images by the proposed and DQN methods. The proposed method provides better results than the DQN method on difficult images having dark areas, small objects, and complex boundaries.

TABLE II
COMPARISON RESULTS OF THE PROPOSED ALGORITHM AND DQN[6].

| method | edge evaluation model input | reward type (Table I) | reward average | no threshold reversal ratio(%) |
|---|---|---|---|---|
| DQN[6] | edge only | case 1 | 0.782 | 100.0 |
| | original image and edge image | case 1 | 0.794 | 100.0 |
| | | case 2 | 0.781 | 100.0 |
| | | case 3 | 0.810 | 100.0 |
| | | case 4 | 0.804 | 100.0 |
| Proposed | | case 1 | 0.815 | 87.1 |
| | | case 2 | 0.832 | 94.9 |
| | | case 3 | 0.837 | 96.7 |
| | | case 4 | 0.794 | 97.4 |

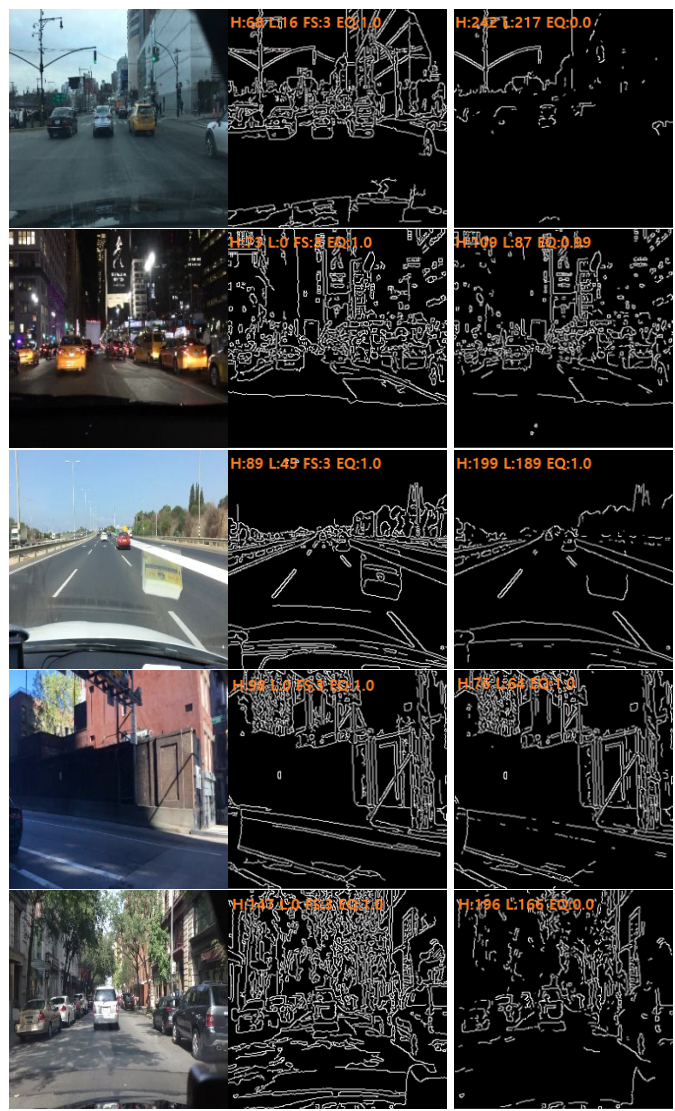

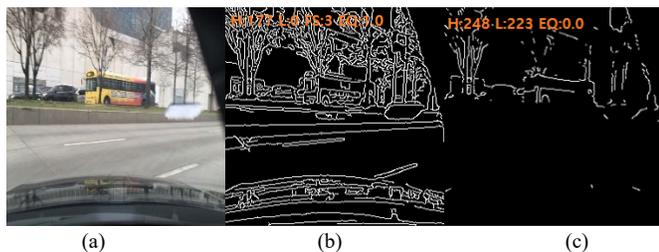

(a)　　　　　(b)　　　　　(c)

Fig. 11. The comparison result of edge image by the proposed algorithm and DQN[6] (a) original image (b) edge image by the proposed algorithm (c) edge image by DQN[6] (d) edge image by DQN[6] with improved evaluation model applied (EQ: stand for edge quality)

Fig. 12 shows the distribution of high and low threshold values selected on 1,000 images by the proposed and DQN methods [6]. We can notice that the proposed method gives a broader distribution of threshold values than the DQN method, as shown in Fig. 12. We can conclude that the proposed algorithm selects appropriate threshold values for images from different natural objects, driving roads, parking lots, day and night, etc.

The edge evaluation model of Fig. 5 shows correct results for most images after training. However, it gives high scores for a bad edge image in some cases. Fig. 13 shows comparison results of edge images of wrong cases by edge evaluation model and the proposed algorithm. Fig. 13(a) is an original image. Fig. 13(b) is an edge image that gives a wrong score of the edge evaluation model. Fig. 13(c) is the edge by the proposed algorithm using an image of Fig. 13(a). We can notice that the proposed algorithm gives improved results, although the edge evaluation model has some drawbacks. It shows that training is possible even using an edge evaluation model with low accuracy, and ultimately it provides improved results.

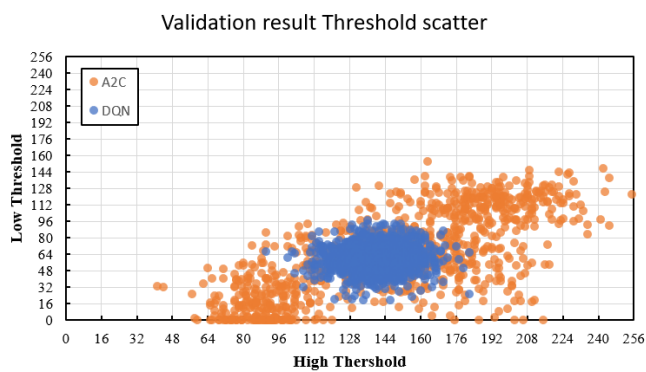

Fig. 12. The distribution of high and low thresholds by the proposed algorithm and DQN[6].

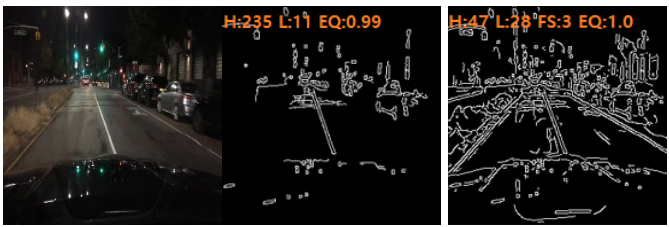

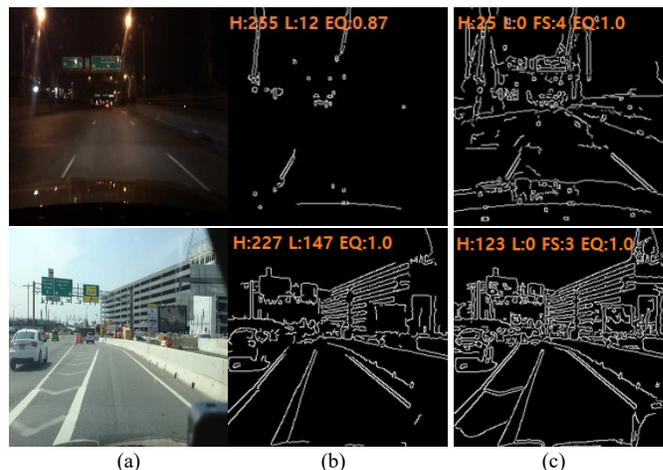

(a)　　　　　(b)　　　　　(c)

Fig 13. Result of the proposed algorithm on images having false edge evaluation results (a) original image (b) false cases by edge evaluation model (c) result by the proposed algorithm.

### C. Experimental Result using Unseen Images

We apply the proposed algorithm on images different from training to evaluate the proposed method's generalization ability qualitatively. Table 3 shows the results applied to YTVOS[42], an environment different from the environment used for training, and the results are from 1,070 images in YTVOS. In the BDD100K image used for the training, case 3, which provides the best result in the proposed model, showed an average reward value of 0.837, and when applied to the YTVOS image, the average reward value was 0.746. The average reward value shows a lower value in a new environment in all cases. The proposed method provides improved results compared to DQN [6] with improved evaluation model applied.

Fig. 14 shows the results of the proposed method applied to various images of YTVOS. Table IV shows the results when additional learning was performed using about 14,000 images of an environment different from the training environment. In this case, it can be seen that there is no significant difference from the existing results.

In Table III and Table IV, both the proposed method and the DQN[6] method used the model shown in Fig. 5 as the edge evaluation model, which uses the original and edge images as inputs. We obtained the statistics in Table III and Table IV using 1,070 images.

For the compensation model, case 3 in Table 1 provides the best results in DQN and the proposed method. The DQN method shows an improvement of 4.1% from 0.721 to 0.751, while the proposed method shows a gain of 7.4% from 0.746 to 0.801 when we do additional training using images of a new environment,

TABLE III. COMPARISON RESULT WHEN APPLYING ON UNSEEN IMAGES OF YTVOS[42].

| method | reward type (Table I) | reward average | no threshold reversal ratio(%) |
|---|---|---|---|
| DQN[6] | case 1 | 0.658 | 100.0 |
|  | case 2 | 0.684 | 100.0 |
|  | case 3 | 0.721 | 100.0 |
|  | case 4 | 0.699 | 100.0 |
| Proposed | case 1 | 0.715 | 86.6 |

| | case 2 | 0.723 | 94.7 |
| | case 3 | 0.746 | 95.8 |
| | case 4 | 0.698 | 88.6 |

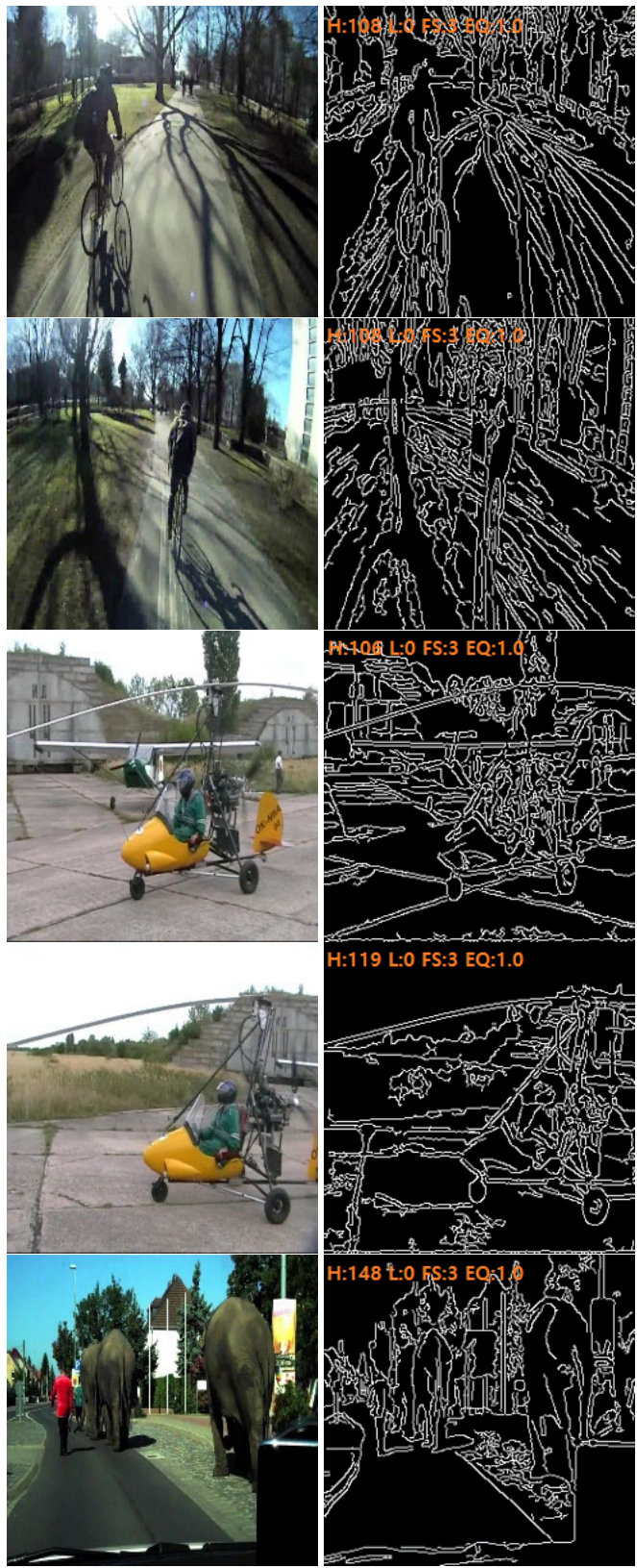

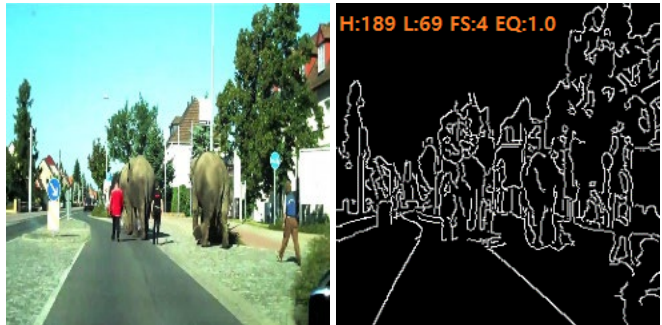

Fig. 14. Results of the proposed algorithm on the unseen images of YTVOS[42].

TABLE IV
RESULT OF THE PROPOSED ALGORITHM AFTER RETRAINING USING YTVOS[42] IMAGES.

| method | reward type (Table I) | reward average | relative improvement(%) | no threshold reversal ratio(%) |
|---|---|---|---|---|
| DQN[6] | case 1 | 0.679 | 3.19 | 100.0 |
| | case 2 | 0.671 | -1.90 | 100.0 |
| | case 3 | 0.752 | 4.30 | 100.0 |
| | case 4 | 0.634 | -9.30 | 100.0 |
| proposed | case 1 | 0.735 | 2.80 | 89.0 |
| | case 2 | 0.779 | 7.75 | 96.5 |
| | case 3 | 0.801 | 7.37 | 97.2 |
| | case 4 | 0.657 | -5.87 | 86.2 |

## V. CONCLUSION

This paper proposes an algorithm for automatically selecting suitable three threshold values for the Canny edge detection. We adopt the continuous A2C algorithm to solve the given problem. We propose an actor and critic network and reward model conFig.d using an edge evaluation network.

In the Canny method, the resulting edge images have a similar tendency if their threshold values are similar. We adopted the continuous A2C that produces a normal distribution for three parameters considering this fact. However, there is a disadvantage in that there is a possibility of reversing the high and low threshold values due to the inherent structure of choosing action values from a normal distribution. Nevertheless, the proposed algorithm improves performance compared to an algorithm using the DQN.

The training results confirmed that the limitations of the CNN model used for the edge quality evaluation were improved in the model learned through reinforcement learning. Compared to the existing supervised learning, this shows a lower dependence on the learning data.

On the other hand, in the case of the filter size determined through the model, the results were confirmed to be biased towards a specific value. This was judged to be a limitation caused by the biased use of only data with high object density in the image during training. We intend to conduct further research to solve this problem.